\pdfoutput=1
\documentclass[11pt]{article}

\usepackage{EMNLP2022}

\usepackage{times}
\usepackage{latexsym}

\usepackage[T1]{fontenc}
\usepackage[utf8]{inputenc}

\usepackage{microtype}

\usepackage{algorithm}
\usepackage{algpseudocode}

\usepackage{graphicx}
\usepackage{amsmath}
\usepackage{amsfonts}
\usepackage{array}
\usepackage{comment}
\usepackage{url}
\usepackage{amsmath}
\usepackage{comment}

\usepackage{inconsolata}

\title{MEE: A Novel Multilingual Event Extraction Dataset}

\author{
Amir Pouran Ben Veyseh\textsuperscript{\rm 1}, Javid Ebrahimi\textsuperscript{\rm 1}, \\ 
{\bf Franck Dernoncourt\textsuperscript{\rm 2}, and}
{\bf Thien Huu Nguyen\textsuperscript{\rm 1}} \\
\textsuperscript{\rm 1}Department of Computer Science, University of Oregon, Eugene, OR, USA\\
\textsuperscript{\rm 2}Adobe Research, Seattle, WA, USA\\
{\tt \{apouranb, thien\}@cs.uoregon.edu; jebivid@gmail.com} \\
{\tt franck.dernoncourt@adobe.com}
}

\begin{document}
\maketitle
\begin{abstract}
Event Extraction (EE) is one of the fundamental tasks in Information Extraction (IE) that aims to recognize event mentions and their arguments (i.e., participants) from text. Due to its importance, extensive methods and resources have been developed for Event Extraction. However, one limitation of current research for EE involves the under-exploration for non-English languages in which the lack of high-quality multilingual EE datasets for model training and evaluation has been the main hindrance. To address this limitation, we propose a novel Multilingual Event Extraction dataset (MEE) that provides annotation for more than 50K event mentions in 8 typologically different languages. MEE comprehensively annotates data for entity mentions, event triggers and event arguments. We conduct extensive experiments on the proposed dataset to reveal challenges and opportunities for multilingual EE. %To foster future research in this direction, our dataset will be publicly available.

%Event Extraction (EE) is one of the fundamental tasks in Information Extraction (IE). In EE, the objective is to recognize the mentions of events, i.e., change of state of real world entities, and also to identify the participants and attributes of each event mention, i.e., event arguments. Due to its importance, several methods and resources have been proposed for Event Extraction. However, one of the limitations in the current research is that EE is less-explored in non-English languages. One of the reasons for lack of research for EE in other languages is the shortcoming of a high quality multilingual EE dataset that provide enough data for training and evaluation of EE in different languages. To address this limitation, in this work, we propose MEE (Multilingual Event Extraction) dataset which provide more than 50K event mention annotations in 8 typologically different languages. MEE provide annotations for entity mentions, event triggers and event arguments. We conduct extensive experiments on the proposed dataset, revealing the challenging nature of the multilingual event extraction task. To foster research in this direction, the dataset will be publicly available. 
\end{abstract}

\section{Introduction}

Event Extraction (EE) is one of the major tasks of Information Extraction (IE) for text. In a complete EE pipeline, three major goals should be pursued: (1) Entity Mention Detection (EMD): to recognize mentions of real world entities; (2) Event Detection (ED): to identify event mentions/triggers and their types. An event trigger is a word or phrase that most clearly refers to the occurrence of an event; and (3) Event Argument Extraction (EAE): to find participants/arguments of an event mentioned in text. A participant is an entity mention that has an specific role in a given event mention. For instance, in the sentence ``\textit{The soldiers were hit by the forces.}'', there are two entity mentions ``\textit{soldiers}'' and ``\textit{forces}'' of types {\it PERSON} and {\it ORGANIZATION} and an event trigger ``\textit{hit}'' of type \textit{ATTACK}. Also, the two event mentions ``\textit{soldiers}'' and ``\textit{forces}'' play the argument roles of \textit{Victim} and \textit{Attacker} (respectively) in the {\it ATTACK} event. An EE system could be employed in other downstream applications such as Question Answering, Knowledge Base Population and Text Summarization to assist extracting information about events in text. 

%Prior work for EE can also be categorized into pipeline and joint inference approaches. In the pipeline approach, an ED model is first used to extract event triggers that will be further sent into a separate EAE model to identify participates for event arguments \cite{ahn2006stages,chen2015event}. In contrast, a joint inference model for EE attempts to simultaneously predict event triggers and arguments in a single process to avoid error propagation \cite{li2013joint,nguyen2016joint}.

Multiple methods have been proposed for Event Extraction. Early work has employed feature-based models \cite{ahn2006stages,ji2008refining,liao2010filtered,hong2011using,li2013joint,yang2016joint} while later methods have explored deep learning to present state-of-the-art performance for Event Extraction \cite{nguyen2015event,chen2015event,nguyen2016joint,sha2018jointly,wang2019adversarial,lai2020event,pouran-ben-veyseh-etal-2020-graph,lin2020oneie,nguyen2021cross,liu-etal-2022-dynamic}. However, despite all advancements on event extraction in recent years, a major limitation of current EE research is to overly focus on a few popular languages, thus failing to adequately reveal challenges and generalization of models in many other languages of the world. As such, a critical barrier for studying EE over multiple languages is the lack of high quality datasets that fully annotate data for many other languages for EE. For instance, the most popular dataset for EE, i.e., ACE 2005 \cite{walker05ace}, only provide annotations for three languages English, Chinese and Arabic while TAC KBP datasets \cite{mitamura16overview,mitamura17event} only supports English, Chinese and Spanish. The TempEval-2 dataset \cite{verhagen2010semeval} involves 6 languages; however, it does not offer event argument annotation. Even worse, recently created datasets, e.g., MAVEN \cite{wang2020maven}, RAMS \cite{ebner2020multi}, and WikiEvents \cite{li-etal-2021-document}, are only annotated for English. In all, such language and task limitations prevents research to comprehensively develop and evaluate EE methods over different languages and multilingual settings. Moreover, the limited size of these datasets, i.e. less than 11K and 27K in ACE 2005 and TempEval-2 respectively, hinders training of data-hungry deep learning models. Finally, we note that important multilingual datasets for EE, e.g., ACE 2005 and TAC KBP, are not publicly available, which further restricts research on this domain.  

%there is a major limitation in the current research. Specifically, the majority of the prior works are limited to the English language. Therefore, the effectiveness of the current methods is not studied in other languages. One of the importance barriers for studying EE in other languages is the lack of high quality dataset that covers other languages. Specifically, the existing datasets, e.g., ACE 2005 \cite{walker05ace}, Tack KBP \cite{mitamura16overview,mitamura17event} or TempEval-2 \cite{verhagen2010semeval}, provide annotations for limited number of languages, i.e., 3 languages for ACE 2005 and TAC KBP and 6 languages for TempEval-2. This limitation restrics any research for multilingual or cross-lingual EE models. Moreover, the limited size of these datasets, i.e. less than 11K and 27K in ACE 2005 and TempEval-2 respectively, hinders training of data-hungry deep learning models. Finally, some the existing resources are not publicly available, e.g., ACE 2005, which further restricts research on this domain.  

To address these limitations, in this work, we propose a large-scale Multilingual Event Extraction (MEE) dataset that covers 8 typologically different languages from multiple language families, including English, Spanish, Portuguese, Polish, Turkish, Hindi, Korean, and Japanese. As such, Portuguese, Polish, Turkish, Hindi, and Japanese are not explored in the popular multilingual datasets for EE, i.e., ACE 2005 and TAC KBP. Importantly, to enable public data sharing and diversity the data, we employ Wikipedia articles for the 8 languages in diverse topics (i.e., Economy, Politics, Technology, Crime, Nature and Military) for EE annotation.

Our dataset comprehensively annotates each document in a language for all the three sub-tasks EMD, ED, and EAE. To be consistent with prior EE research, we inherit the type anthologies for such tasks from the ACE 2005 dataset that provides well-designed guidelines and examples for the types. In particular, we include 7 entity types, 8 event types and 16 event sub-types, along with 23 argument roles in MEE to facilitate EE annotation over multiple languages. Overall, our dataset involves more than 415K entity mentions, 50K event triggers, and 38K arguments, which are much larger than previous multilingual EE datasets to better support model training and evaluation with deep learning.

Due to shared information schema over all the languages, our MEE dataset enables cross-lingual transfer learning evaluation of MEE models where training and test data comes from different languages. To this end, we conduct comprehensive experiments for both monolingual and cross-lingual learning settings to provide insights for language-specific challenges and cross-lingual generalization of EE methods. By examining both pipeline and joint inference models for EE, our experiments show that the proposed dataset present unique challenges with less satisfactory performance of existing EE models, especially for cross-lingual settings, thus calling for more research efforts for multilingual EE in the future.

%We will publicly release our new dataset to foster future research in this area.

%Compared to the existing resources for Event Extraction, MEE provide unique challenges which foster future research on EE. First, MEE employs a unified schema across 8 different languages which can be used to study the language specific challenges and also to compare the performance of the models across multiple language families. Also, since MEE employs ACE 2005 ontology, prior works employed trained on ACE 2005 dataset can also be compared against MEE. Second, compared to the existing EE datasets, MEE provide more training samples which can be used to train and evaluate deep models with more parameters. For instance, while RAMS \cite{ebner2020multi} dataset provide annotations for 9K examples with 21K arguments, MEE has 31K examples with more than 38K event argument extractions. 

%To demonstrate the challenges introduced by MEE, in this work, we conduct comprehensive analysis in monolingual and cross-lingual settings using both joint and pipeline baselines. Our experiments show that the proposed dataset is more challenging than the existing resources and more research is required, specifically for the cross-lingual setting. 

\begin{table*}[]
    \centering
    \resizebox{.9\textwidth}{!}{
    \begin{tabular}{l|cccccccc}
        Category & English & Portuguese & Spanish & Polish & Turkish & Hindi & Japanese & Korean \\ \hline
        Economy & 1,095 & 112 & 168 & 315 & 297 & 189 & 199 & 250 \\
        Politics & 3,202 & 308 & 772 & 1,270 & 1,233 & 349 & 232 & 248 \\
        Technology & 2,171 & 189 & 400 & 712 & 815 & 295 & 312 & 249 \\
        Crimes & 893 & 78 & 220 & 152 & 118 & 95 & 80 & 73 \\
        Nature & 1,195 & 398 & 705 & 455 & 398 & 245 & 299 & 185 \\ 
        Military & 4,444 & 415 & 1,003 & 1,575 & 1,619 & 326 & 378 & 495 \\ \hline
        Total & 13,000 & 1,500 & 3,268 & 4,479 & 4,480 & 1,499 & 1,500 & 1,500 \\
    \end{tabular}
    }
    \caption{Numbers of annotated segments in each Wikipedia subcategory for our 8 languages.}
    \label{tab:topic_stats}
\end{table*}

\section{Data Annotation}

We follow the entity/event type definition and annotation guidelines from the popular ACE 2005 dataset to benefit from its well-designed documentation and be consistent with prior EE research. As such, entity mentions refer to mentions of real-world entities in text that can be expressed via names, nominals, and pronouns. Entity Mention Extraction (EMD) is more general than Named Entity Recognition that only concerns names of entities. In addition, an event is defined as an incident whose occurrence changes the state of real world entities. An event mention is the part of input text that refers to an event that consists of two components: (1) Event Trigger: the words that most clearly refer to the occurrence of the event. It is noteworthy that we allow an event trigger to span multiple words to accommodate trigger annotation for multiple languages. For instance, in the Turkish phrase ``\textit{tayin etmek}'', both words are necessary to indicate an event trigger of type ``\textit{Appoint}";  and (2) Event Arguments: the entity mentions that are involved in the event with some roles.

%In the proposed dataset MEE, we follow the event definition that is provided by ACE 2005 \cite{walker05ace}. Specifically, an event is defined as an incidents that its occurrence changes the state of the real world entities. An event mention is part of the text that refers to the event. An event mention consists of two components: (1) Event Trigger: The word(s) tha most clearly refer to the occurrence of the event. It is note worthy that to accommodate the requirements of all languages, we allow an event trigger to span multiple words. For instance, in the Turkish phrase ``\textit{tayin etmek}" both words are annotated as event trigger for the event of ``\textit{Appoint}";  and (2) Event Arguments: The mentions of the entities that are involved in the event. Also, the attributes, e.g., time, of the event are considered as the event argument. Also, since the event arguments are selected from entity mentions, prior to event extraction, we annotate the corpus for named entity recognition (NER) task. To be consistent with the prior works and also to benefit from the well-documented annotation guideline of ACE 2005, we follow the same instructions as ACE 2005 for event extraction and named entity recognition annotation of MEE.

Based on the ACE 2005 dataset, our dataset annotates entity mentions for 7 entity types: \textbf{PERSON} (human entities), \textbf{ORGANIZATION} (corporations, agencies, and other groups of people), \textbf{GPE} (geographical regions defined by political and/or social groups), \textbf{LOCATION} (geographical entities such as landmasses or bodies of water), \textbf{FACILITY} (buildings and other permanent man-made structures), \textbf{VEHICLE} (physical devices primarily designed to move an object from one location to another), and \textbf{WEAPON} (physical devices primarily used as instruments for physically harming). For event types, to avoid confusion and improve data quality, we prune the original ACE 2005 event types to only include the types that are not ambiguous across multiple languages. For instance, in Turkish, the event types \textit{Sentence} and \textit{Convict} are very similar (both can be evoked by the phrase ``\textit{Mahkum etmek}'') so they are not retained in our dataset. As such, we preserve 8 event types and 16 sub-types that are distinct enough for annotation in our dataset. Finally, for event arguments, we preserve all 23 argument roles in the ACE 2005 dataset. Table \ref{tab:event_schema} shows the list of event types along with their argument roles in our dataset.

\subsection{Data Preparation}

Our dataset MEE covers 8 different languages, i.e., English, Spanish, Portuguese, Polish, Turkish, Hindi, Korean and Japanese. These languages are selected based on their diversity in terms of typology and their novelty with respect to existing multilingual EE datasets. For each language, we employ its latest dump of Wikipedia articles as raw data for annotation. To focus on event data, we select articles in the sub-categories under category {\it Event} in Wikipedia. In particular, the following sub-categories are considered to improve topic diversity: Economy, Politics, Technology, Crimes, Nature, and Military. Note that we start with these categories in English Wikipedia. Afterward, we follow interlinks between the categories in different languages to locate the intended categories for Wikipedia for non-English languages in MEE.

We process the collected articles with the WikiExtractor tool \cite{Wikiextractor2015} to obtain clean textual data and meta-data for each article. The textual data is then split into sentences and tokenized into words by the multilingual NLP toolkit Trankit \cite{nguyen2021trankit}. Afterward, to annotate the data with entity and event mentions, one approach is to directly ask annotators to read each article entirely for annotation. However, as the articles in Wikipedia might be lengthy, this approach can be overwhelming for annotators, thus hindering their attention and lowering quality of annotated data. To address this issue, we follow prior dataset creation efforts for EE, i.e., RAMS \cite{ebner2020multi}, to divide the articles into segments of five consecutive sentences. Each segment will then be annotated separately for EE tasks so annotators can better capture the entire context to provide entity and event annotation. Note that similar to RAMS, we annotate all event arguments in a text segment for each event trigger, thus allowing event arguments to appear in different sentences from the event trigger (i.e., document-level EAE). Finally, to accomodate our budget, a sample of text segments is obtained for each language for annotation. The numbers of selected text segments for each category per language in our dataset are presented in Table \ref{tab:topic_stats}.

%This decision is also supported in other related prior works to deliver high extraction performance for the event types in ACE 2005 \cite{nguyen2015event,Nguyen:18a,wang2019adversarial,yang2019exploring,cui2020edge}, including models for multiple languages \cite{mhamdi2019contextualized,ahmad2020gate,nguyen-etal-2021-crosslingual}.

%To annotate the collected articles with entity and event mentions, one approach is to directly ask the annotators to read the entire article and annotate it. However, since some of the articles might be lengthy, this approach could lower the quality of the data as the annotators might miss some of the mentions in the entire article. To address this issue, we follow the prior work, i.e., RAMS \cite{ebner2020multi}, to segment the documents into sections of 5 sentences. Each section will be annotated separately, therefore the annotator can remember the entire context provided to them while annotating the entities and events. This decision is also supported in other related prior works to deliver high extraction performance for the event types in ACE 2005 \cite{nguyen2015event,Nguyen:18a,wang2019adversarial,yang2019exploring,cui2020edge}, including models for multiple languages \cite{mhamdi2019contextualized,ahmad2020gate,nguyen-etal-2021-crosslingual}.

\begin{table*}[]
    \centering
    \resizebox{.98\textwidth}{!}{
    \begin{tabular}{c|cccccccc}
        Language & \#Seg. & Avg. Length & \#Entities & \#Triggers & \#Arguments & Challenging Entity Type & Challenging Trigger Type & Language Family \\ \hline
        English & 13,000 & 123 & 190,592 & 17,642 & 13,548 & GPE & Personnel & Germanic \\
        Spanish & 3,268 & 112 & 48,001 & 6,064 & 802 & GPE & Conflict & Italic \\
        Portuguese & 1,500 & 102 & 25,463 & 1,953 & 12,329 & Location & Personnel & Italic \\
        Polish & 4,479 & 108 & 62,971 & 10,875 & 3,395 & Facility & Transaction & Balto-Slavic \\
        Turkish & 4,480 & 117 & 38,469 & 8,390 & 1,416 & GPE & Personnel & Turkic \\
        Hindi & 1,499 & 98 & 18,797 & 1,810 & 2,117 & Facility & Conflict & Indo-Iranian \\
        Japanese & 1,500 & 99 & 19,174 & 2,152 & 3,399 & Location & Personnel & Japonic \\
        Korean & 1,500 & 103 & 12,508 & 1,125 & 1,742 & GPE & Personnel & Koreanic \\ \hline\
        Total (MEE) & 31,226 & - & 415,975 & 50,011 & 38,748 & - & - & - \\ \hline
        %ACE 2005 & 599 & 394 & 54,824 & 5,349 & 54,020 & - & - & - \\
        %TAC KBP & 360 & 660 & - & 6,538 & - & - & - & - \\
        %TempEval-2 & 5,880 & 30 & - & 27,007 & - & - & - & -
    \end{tabular}
    }
    \caption{Statistics of the MEE dataset. \#Seg. represents the numbers of annotated text segments for each language. All annotated segments consist of 5 sentences and their lengths (Avg. Length) are computed in terms of numbers of tokens. ``\textit{Challenging Type}'' indicates the types where entity or event trigger annotation involves largest disagreement between annotators in each language.
    }
    %and comparable prior multilingual EE datasets
    \label{tab:data_stats}
\end{table*}

\begin{table}[]
    \centering
    \resizebox{.45\textwidth}{!}{
    \begin{tabular}{l|cccc}
        Language & \#Annotator & EMD & ED & EAE \\ \hline
        English & 10 & 0.792 & 0.834 & 0.820 \\
        Spanish & 10 & 0.788 & 0.812 & 0.823 \\
        Portuguese & 5 & 0.791 & 0.803 & 0.799 \\
        Polish & 8 & 0.780 & 0.799 & 0.813 \\
        Turkish & 10 & 0.785 & 0.813 & 0.822 \\
        Hindi & 6 & 0.790 & 0.803 & 0.812 \\
        Japanese & 5 & 0.793 & 0.789 & 0.780 \\
        Korean & 6 & 0.802 & 0.810 & 0.825 \\
    \end{tabular}
    }
    \caption{Number of annotators and agreement scores for 8 languages in MEE for Entity Mention Detection (EMD), Event Detection (ED) and Event Argument Extraction (EAE).}
    \label{tab:agreement}
\end{table}

%%%%\begin{table}[]
%%%%    \centering
%%%%    \resizebox{.45\textwidth}{!}{
%%%%    \begin{tabular}{l|c|c}
%%%%        Language & Entity & Trigger \\ \hline
%%%%        English & Person, GPE, Organization & Life, Conflict, Movement \\
%%%%        Spanish & Person, GPE, Organization & Personnel, Life, Conflict \\
%%%%        Portuguese & Person, GPE, Organization & Life, Movement, Conflict \\
%%%%        Polish & Person, GPE, Facility & Life, Personnel, Conflict \\
%%%%        Turkish & Person, GPE, Organization & Life, Conflict, Personnel \\
%%%%        Hindi & Person, GPE, Facility & Life, Movement, Conflict \\
%%%%        Japanese & Person, Organization, GPE & Personnel, Life, Conflict \\
%%%%        Korean & Person, GPE, Organization & Personnel, Life, Conflict
%%%%    \end{tabular}
%%%%    }
%%%%    \caption{Frequent Entity and Trigger Types in MEE}
%%%%    \label{tab:frequency}
%%%%\end{table}

\subsection{Annotation Process}

To annotate the sampled article segments, we employ the crowd-sourcing platform \url{upwork.com} that allows us to hire freelancers across the globe with different expertise. For each language in our dataset, we choose native speakers as annotator candidates. In addition, we require them to be fluent in English, have experience in related tasks (i.e., data annotation for information extraction), and have approval rate higher than 95\% (i.e., provided in their profiles). The candidates are first provided with annotation guidelines and interfaces in English. Afterward, they are invited to an annotation test for entity mentions, event triggers, and arguments. Those candidates who correctly annotate all test cases are then officially hired to work on our annotation jobs. Table \ref{tab:agreement} shows the numbers of annotators who are hired to annotate data for each language in our dataset. Next, before the actual annotation process, the English annotation guideline and examples are translated to each target language by the hired annotators. Any language-specific confusions and rules for annotation is discussed and included in the translation to create a common understanding. Finally, our language experts will review the annotation guideline in each language to avoid conflicts across languages to be used for actual annotation.

%To annotate the collected article segments, we employ the crowd-sourcing platform \url{upwork.com}. This platform provide opportunity to hire freelancers across the globe with different expertise. In our work, we hire annotators that are native in the target language, have also experience in related task, i.e., data annotation for information extraction, and have approval rate higher than 95\%, which are provided in their profile. To hire the annotators, we first require them to pass an entity and event annotation test. The test is conducted in English and also the target language. A Language expert verifies the performance of the annotators. Those freelancers that correctly annotate all test cases are hired to work on entire corpus. The Table \ref{tab:agreement} shows the number of annotators that are hired for each language. Before the actual annotation process, the guideline which is originally written in English is translated to the target language by the hired annotators. The conflicts and confusions are discussed with language experts in the target language. The final version of each annotation guideline for each language is again compared with the English version to encourage the same annotation policies across all languages.

Our annotation process is done in three separate steps to annotate data for three EE tasks with entity mentions, event triggers, and event arguments in this order. In particular, the annotation for a later task will be performed over the text segments that have been annotated and finalized for previous tasks (e.g., event arguments will be annotated over segments that are already provided with entity mentions and event triggers). As such, for each task, 20\% of text segments for each language will be co-annotated by the annotators to measure agreement score. The remaining 80\% of text segments will be distributed and annotated separately by the annotators for each language. Based on the Krippendorff’s alpha \cite{krippendorff2011computing} with MASI distance metric \cite{passonneau2006measuring}, we report the inter-annotator agreements (IAA) for each task and language in Table \ref{tab:agreement}, showing high agreement scores and quality of our MEE dataset. Note that after independent annotation for each EE task, the annotators also share their annotations and communicate with each other to resolve any conflicts and finalize our data.

\begin{table*}[]
    \centering
    \begin{tabular}{c|l|l}
        \textbf{ID} & \textbf{Event} & \textbf{Arguments} \\ \hline \hline
        1 & Life\_Be-Born & Person, Time, Place \\\hline
        2 & Life\_Marry & Person, Time, Place \\\hline
        3 & Life\_Divorce & Person, Time, Place \\\hline
        4 & Life\_Injure & Agent, Victim, Instrument, Time, Place \\\hline
        5 & Life\_Die & Agent, Victim, Instrument, Time, Place \\\hline
        6 & Movement\_Transport & Agent, Artifact, Vehicle, Price, Origin, Destination, Time \\\hline
        7 & Transaction\_Transfer-Ownership & Buyer, Seller, Beneficiary, Price, Artifact, Time, Place \\\hline
        8 & Transaction\_Transfer-Money & Giver, Recipient, Beneficiary, Money, Time, Place \\\hline
        9 & Business\_Start-Organization & Agent, Organization, Time, Place \\\hline
        10 & Conflict\_Attack & Attacker, Target, Instrument, Time, Place \\\hline
        11 & Conflict\_Attack & Entity, Time, Place \\\hline
        12 & Contact\_Meet & Entity, Time, Place \\\hline
        13 & Contact\_Phone-Write & Entity, Time \\\hline
        14 & Personnel\_Start-Position & Person, Entity, Position, Time, Place \\\hline
        15 & Personnel\_End-Position & Person, Entity, Position, Time, Place \\\hline
        16 & Justice\_Arrest-Jail & Person, Agent, Crime, Time, Place
    \end{tabular}
    \caption{Event types and argument roles for each type in MEE. The types and roles are inherited from the event extraction annotation guideline in the ACE 2005 dataset \cite{walker05ace}.}
    \label{tab:event_schema}
\end{table*}

%To annotate the data, we divide the entire article segments for each language into to splits. Specifically, 80\% of the article segments, which are randomly selected, are distributed among annotators of each language to increase the annotation efficiency. The rest, i.e., 20\% of the segments, are shared across all annotators of the target language to measure the agreement between annotators. Finally, to resolve the conflicts, the annotators share their annotations and communicate with each other. In order to asses the quality of the annotation, we measure the inter-annotator agreements (IAA) using Krippendorff’s alpha \cite{krippendorff2011computing} with MASI distance metric \cite{passonneau2006measuring}. The results are shown in Table \ref{tab:agreement}. This table shows that there is high agreement between annotators across all 8 languages. This high agreement implies the quality of the prepared dataset. It is note worthy that among all three tasks of NER, ED and EAE, the NER annotation has the lowest agreement which indicates the high degree of ambiguity for this task compared to the others. 

\subsection{Data Analysis}

%The proposed dataset contains more than 415K entity mentions, 50K event mentions and 38K argument mentions. The sheer amount of samples for each task provide more opportunity for future research to train and evaluate deep models with high number of parameters.

%Specifically, considering the segment length, while all languages have segments of 5-sentence long, some of them have more words than others, e.g., 123 words/segment in English compared to 98 words/segment in Hindi. The different length of the sentences in these languages indicates the inherent difference between these languages for providing context in a sentence which can be challenging for a multilingual or cross-lingual model.

Table \ref{tab:data_stats} shows the main statistics of MEE for each language. As such, comparing to the popular multilingual ACE 2005 dataset \cite{walker05ace} for EE, our MEE dataset provides more languages (i.e., 3 vs. 8) and much more event mentions (i.e., 11K vs. 50K). For other multilingual datasets for EE, i.e., TAC KBP (with three languages and 6.5K event mentions) \cite{mitamura16overview,mitamura17event} and TempEval-2 (with 6 languages and 27K event mentions) \cite{verhagen2010semeval}, they do not annotate entity mentions and event arguments. In contrast, our MEE dataset fully annotates texts for three EE tasks (i.e., EMD, ED, and EAE) and also with more languages and event mentions. This clearly demonstrates the advantages of our dataset over existing multilingual datasets for EE.

\begin{figure}
    \centering
    \resizebox{.48\textwidth}{!}{
    \includegraphics{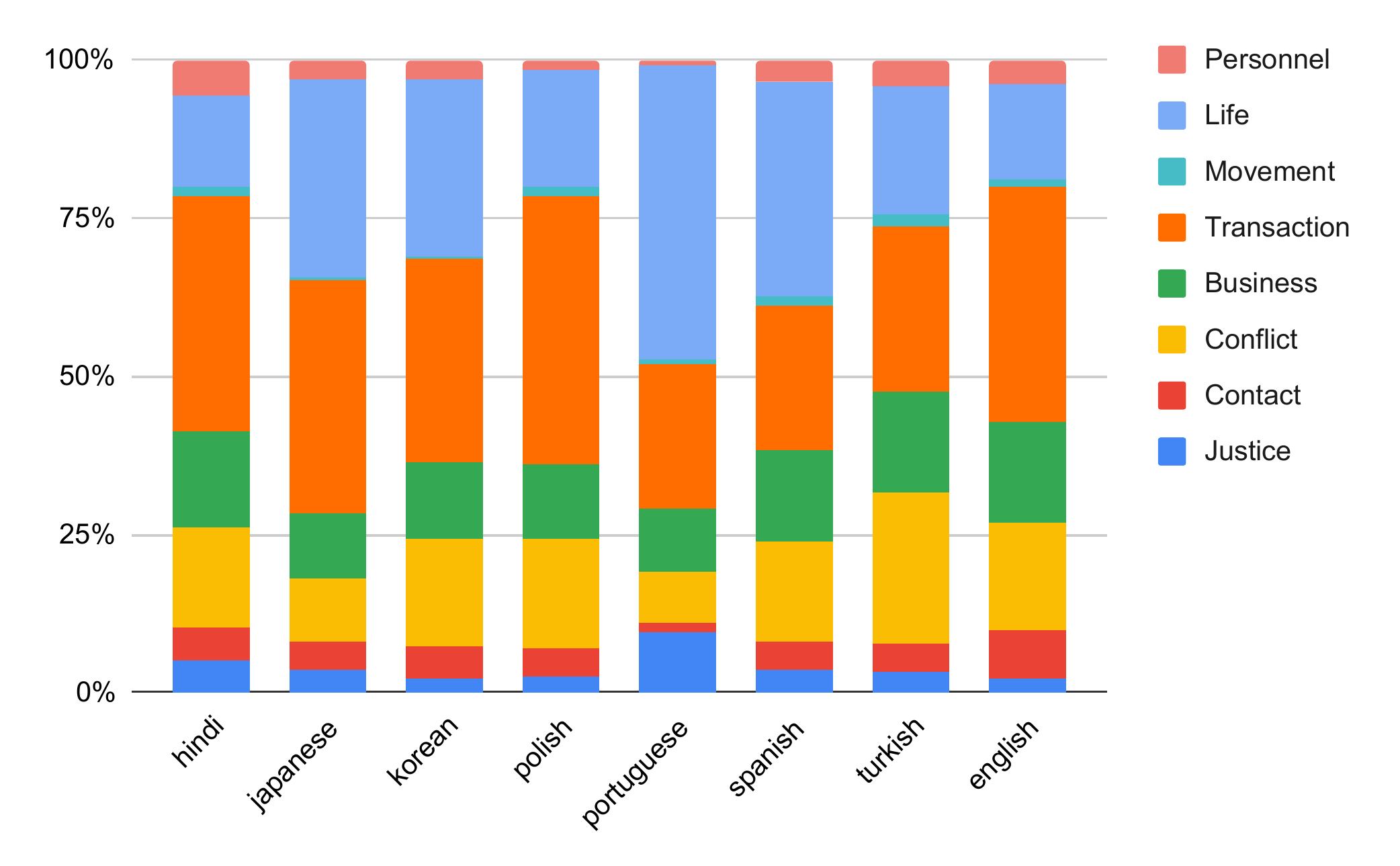}
    }
    \caption{Distributions of event types in each language.}
    \label{fig:event_type}
\end{figure}

\begin{figure}
    \centering
    \resizebox{.48\textwidth}{!}{
    \includegraphics{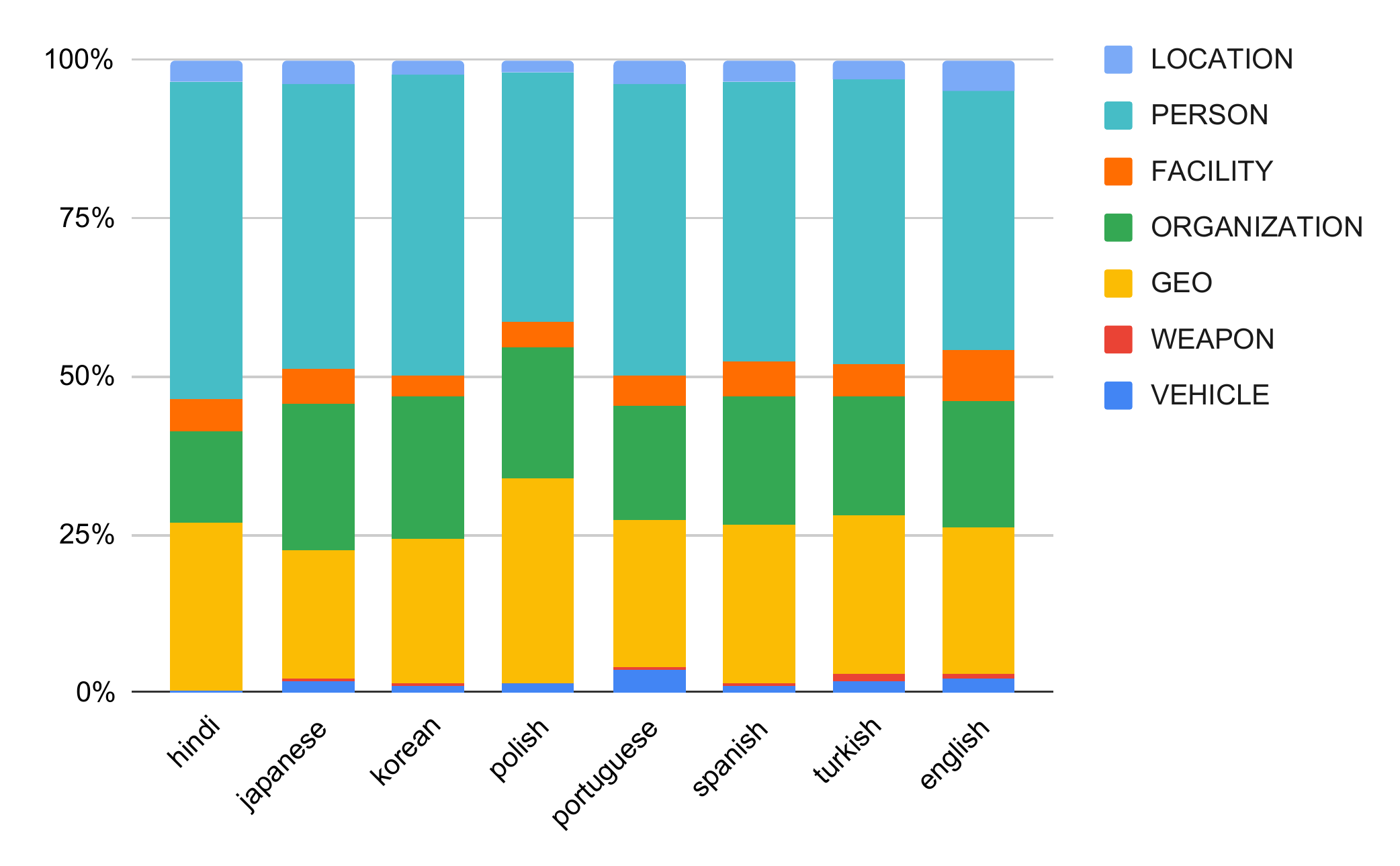}
    }
    \caption{Distributions of entity types in each language.}
    \label{fig:entity_type}
\end{figure}

\begin{figure}
    \centering
    \resizebox{.48\textwidth}{!}{
    \includegraphics{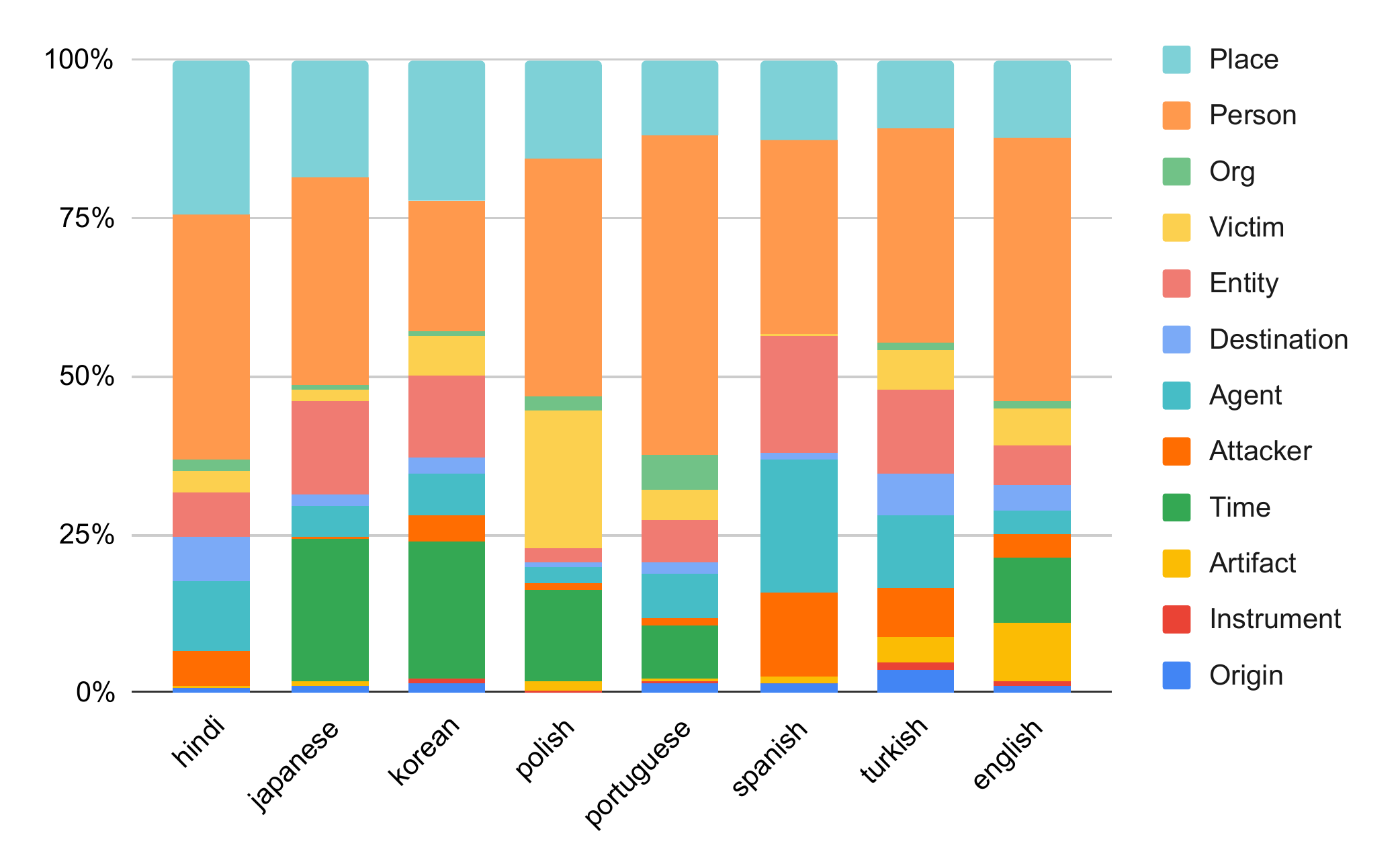}
    }
    \caption{Distributions of most argument roles for each language.}
    \label{fig:argument_type}
\end{figure}

In addition, from the table, we find that the languages in our dataset exhibits diverse densities for entity mentions, event triggers, and arguments in texts. In particular, while the average number of entities in a text segment in Portuguese is 16.9, this number is only 8.3 in Korean. For event density, in Polish, there are 2.4 event mentions per article segment on average while the average number in Korean is only 0.75. Similarly for event arguments, the average number of arguments per event is 6.1 in Portuguese and only 0.75 in English. Further, Table \ref{tab:data_stats} highlights the divergences between languages regarding challenging entity and event types. Specifically, we employ the disagreement rates (i.e., number of disagreements divided by frequency of mentions) between annotators for each entity and event types. Those types that have highest disagreement rates are selected as challenging entity or event types. Finally, Figures \ref{fig:entity_type}, \ref{fig:event_type}, and \ref{fig:argument_type} present the distributions of entity types, event types, and argument roles (respectively) for each language in our dataset, which further demonstrate the differences between languages in MEE. In all, such differences over various dimensions will cause significant challenges for EE models to adapt to new languages (e.g., for cross-lingual transfer learning), thus presenting ample opportunities for multilingual EE research with our dataset.

\begin{table*}[]
    \centering
    \resizebox{.9\textwidth}{!}{
    \begin{tabular}{l|ccc|ccc|ccc}
        Language & \multicolumn{3}{c}{Pipeline} & \multicolumn{3}{c}{OneIE} & \multicolumn{3}{c}{FourIE}  \\ \cline{2-10}
         & Entity & Event & Argument & Entity & Event & Argument & Entity & Event & Argument \\ \hline
         English & 70.32 & 70.58 & 61.14 & 62.18 & 70.09 & 62.94 & 69.72 & 72.19 & 65.89 \\ 
         Spanish & 70.39 & 66.19 & 60.16 & 70.27 & 65.00 & 60.31 & 71.89 & 67.49 & 62.19 \\ 
         Portuguese & 75.13 & 71.33 & 69.15 & 73.19 & 70.13 & 71.27 & 74.98 & 72.99 & 70.17 \\
         Polish & 69.27 & 59.12 & 60.09 & 60.09 & 59.44 & 60.14 & 68.23 & 60.98 & 61.32 \\
         Turkish & 71.88 & 66.09 & 56.19 & 71.98 & 61.27 & 58.72 & 72.33 & 65.13 & 59.80 \\ 
         Hindi & 66.22 & 57.77 & 57.78 & 61.72 & 58.18 & 59.44 & 65.23 & 59.88 & 60.82 \\ 
         Japanese & 68.19 & 67.89 & 68.19 & 71.40 & 65.01 & 63.17 & 70.88 & 66.88 & 70.19 \\ 
         Korean & 57.17 & 61.26 & 67.87 & 55.87 & 61.10 & 65.41 & 58.18 & 60.09 & 69.23 \\ \hline
         Avg. & 68.57 & 65.03 & 62.57 & 65.84 & 63.78 & 62.68 & 68.93 & 65.70 & 64.95 
    \end{tabular}
    }
    \caption{Performance (F1 scores) of models in the monolingual setting using mBERT on MEE.}
    \label{tab:mbert_results}
\end{table*}

\begin{table*}[]
    \centering
    \resizebox{.9\textwidth}{!}{
    \begin{tabular}{l|ccc|ccc|ccc}
        Language & \multicolumn{3}{c}{Pipeline} & \multicolumn{3}{c}{OneIE} & \multicolumn{3}{c}{FourIE}  \\ \cline{2-10}
         & Entity & Event & Argument & Entity & Event & Argument & Entity & Event & Argument \\ \hline
         English & 70.22 & 71.28 & 66.34 & 70.39 & 70.29 & 68.68 & 71.19 & 73.14 & 68.23 \\ 
         Spanish & 70.33 & 64.32 & 61.12 & 70.18 & 62.46 & 62.23 & 72.87 & 65.90 & 63.11 \\ 
         Portuguese & 70.39 & 71.88 & 71.75 & 72.16 & 69.43 & 70.33 & 73.98 & 70.43 & 72.23 \\
         Polish & 69.14 & 60.45 & 61.23 & 72.22 & 63.77 & 60.15 & 70.25 & 62.87 & 62.84 \\
         Turkish & 76.13 & 67.18 & 55.78 & 74.45 & 65.31 & 57.40 & 75.19 & 67.29 & 58.23 \\ 
         Hindi & 65.14 & 59.34 & 58.22 & 61.72 & 58.18 & 59.44 & 66.69 & 61.99 & 62.19 \\ 
         Japanese & 71.34 & 67.77 & 69.19 & 68.20 & 62.89 & 70.90 & 72.82 & 65.27 & 73.55 \\ 
         Korean & 59.13 & 62.34 & 69.70 & 59.99 & 60.55 & 66.89 & 60.24 & 61.18 & 70.09 \\ \hline
         Avg. & 68.98 & 65.57 & 64.17 & 68.84 & 64.36 & 64.57 & 70.40 & 66.01 & 66.31 
    \end{tabular}
    }
    \caption{Performance (F1 scores) of models in the monolingual setting using XLM-RoBERTa on MEE.}
    \label{tab:xlmr_results}
\end{table*}

\section{Experiments}

This section evaluates the state-of-the-art models for Event Extraction to reveal challenges in our new dataset MEE. To this end, the annotated article segments for each language in MEE are randomly split into training/development/test portions with the ratios of 80/10/10. Here, to prevent any information leakage, we ensure that different segments of an article (if any) are only assigned to one portion of the data split for each language. We examine EE models in two different settings: (1) monolingual learning where training and test data of models comes from the same language; (2) cross-lingual transfer learning where models are trained on training data of one language (i.e., the source language), but evaluated directly on test data of the other languages (i.e., the target languages).

%In order to study the challenges of Event Extraction in the proposed MEE dataset, in this section, we evaluate the performance of the state-of-the-art models for entity and event extraction tasks. To this end, the annotated article segments of MEE are randomly split into train/development/test sets with the ratio 80/10/10. It is note worthy that to prevent any information leakage, the segments of the each article are assigned only to one specific split. We evaluate the proposed model in two different settings: (1) Monolingual in which the train set of each languages is employed to train the model, then, it is evaluated on the corresponding test sets for the language; (2) Cross-lingual in which the models are trained on a source language, e.g., English, then, they are evaluated on the test sets of the other languages. 

\noindent {\bf Models}: We evaluate two typical approaches for EE models with pipeline and joint inference in this work. First, for the pipeline approach, a model is trained separately for each of the three tasks in EE, i.e., entity mention detection (EMD), event detection (ED), and event argument extraction (EAE). Here, the EMD and ED tasks are modeled as sequence labeling problems, aiming to predict BIO tag sequences for each input sentence to capture spans and types of entity and event mentions. As such, motivated by previous work \cite{wang2020maven}, our EMD and ED models leverage a pre-trained transformer-based language model to encode the input text. The representation for each token in input text (obtained via average of hidden vectors of word-pieces in the last transformer layer) is then sent into a feed-forward network to compute a tag distribution for the token for training and decoding. For EAE, the task is formulated as a text classification problem in which the input consists of an input text and two word indices for the positions of an event trigger and an entity mention of interest. The goal is to predict the argument role that the entity mention plays for the event. To this end, we also use a pre-trained language model to obtain representations for the tokens in input text. Next, the representations for the event trigger and entity mention words are concatenated and sent to a feed-forward network to predict argument role. Note that the EAE model employs golden entity mentions and event triggers during the training process while the outputs from the EMD and ED models are fed into the EAE model in the test time.

Second, for the joint inference approach, EE models simultaneously predicts entity mentions, event triggers, and arguments in end-to-end fashion to avoid error propagation and leverage inter-dependencies between tasks. To this end, we evaluate two state-of-the-art (SOTA) joint EE models, OneIE \cite{lin2020oneie} and FourIE \cite{nguyen2021cross}, in this work due to their language-agnostic nature. Both OneIE and FourIE utilize pre-trained language models to represent input texts and capture cross-task dependencies for joint inference. Note that these models are original designed to include the relation extraction task between entities. To adapt them to EE, we obtain their implementations from the original papers and remove the relation extraction components. Finally, for performance measure, we report the performance (F1 scores) of EE models over three tasks EMD (Entity), ED (Event), and EAE (Argument) using the same correctness criteria as in prior work \cite{lin2020oneie} (i.e., requiring correct prediction for both offsets and types of entity mentions, event triggers, and argument roles).

\noindent {\bf Hyper-parameters}: To facilitate evaluation with multiple languages, we leverage the multilingual pre-trained language models (PLMs) mBERT \cite{devlin2019bert} and XLM-RoBERTa \cite{conneau2020unsupervised} (base versions) to encode texts for EE models. For the pipeline approach, we fine-tune the hyper-parameters for the EMD, ED, and EAE models over development data for English and apply the selected values for all experiments for consistency. In particular, our hyper-parameters for the pipeline model include: 2 hidden layers with 250 hidden units in each layer for the feed-forward networks, 8 for mini-batch size, and 1$e$-2 for learning rate with the Adam optimizer. For the joint IE models, we utilize the same hyper-parameters suggested in the original papers, i.e., OneIE \cite{lin2020oneie} and FourIE \cite{nguyen2021cross}. 

%Finally, we provide a reproducibility checklist in Appendix \ref{app:repo}.

%In our experiments, we leverage \textbf{mBERT} and \textbf{XLM-RoBERTa} as the multilingual transformer-based encoders for the pipeline and joint models. Also, to fine-tune the hyper-parameters of the pipeline model, we employ the development set of the English split for NER task. Specifically, 2 layers of hidden states with 250 dimensions for the feed-forward networks, mini-batch of size 8, and Adam optimizer with learning rate $1e$-2 are selected. The same hyper-parameters are employed for other languages and tasks. For the joint models, we use the same hyper-parameters as the pre-trained models OneIE \cite{lin2020oneie} and FourIE \cite{nguyen2021cross}.

\begin{table}[]
    \centering
     \resizebox{.48\textwidth}{!}{
    \begin{tabular}{l|ccc|ccc}
        Language & \multicolumn{3}{c}{XLM-RoBERTa} & \multicolumn{3}{c}{mBERT} \\ \cline{2-7}
        & Entity & Event & Argument & Entity & Event & Argument \\ \hline
        English & 69.72 & 72.19 & 65.89 & 71.19 & 73.14 & 68.23 \\ \hline
        Spanish & 61.96 & 59.70 & 52.23 & 60.72 & 60.06 & 50.77 \\
        Portuguese & 59.98 & 54.80 & 52.23 & 56.17 & 52.98 & 50.28 \\
        Polish & 52.89 & 51.78 & 52.44 & 53.44 & 50.29 & 53.56 \\
        Turkish & 60.13 & 53.32 & 52.19 & 59.19 & 52.76 & 53.10 \\
        Hindi & 56.32 & 59.76 & 57.17 & 55.39 & 58.44 & 55.65 \\
        Japanese & 41.13 & 44.95 & 40.13 & 42.43 & 43.76 & 41.18 \\
        Korean & 45.78 & 42.99 & 43.04 & 44.78 & 40.22 & 41.14 \\
    \end{tabular}
    }
    \caption{Cross-lingual performance (F1 scores) of {\bf FourIE} when it is trained on English training data and evaluated on test data of other languages in MEE.}
    \label{tab:crosslingual_results}
\end{table}

\noindent \textbf{Results}: The results for monolingual experiments over different languages in MEE are presented in Tables \ref{tab:mbert_results} and \ref{tab:xlmr_results} (i.e.,  with mBERT and XLM-RoBERTa encoders respectively). There are several observations from the tables. First, the models' performance on individual languages and on average for all three tasks EMD, ED, and EAE is still far from being perfect (i.e., all average performance is less than 69\%), thus indicating considerable challenges in our multilingual EE dataset for future research. In addition, comparing the current state-of-the-art joint IE model (i.e., FourIE) with the pipeline method, we find that FourIE is better than the pipeline model on average, especially for the EAE task with significant performance gap. As such, we attribute this to the ability of joint models to mitigate error propagation to EAE from EMD and ED to boost the performance. Due to its best average performance, FourIE will be leveraged in our next experiments. Finally, we find that XLM-RoBERTa generally has better performance than mBERT (i.e., on average) for EE models. Future research can thus focus on XLM-RoBERTa to develop better EE models for multilingual settings.

\noindent {\bf Cross-lingual Evaluation}: To further understand the cross-lingual generalization challenges in MEE, Table \ref{tab:crosslingual_results} reports the performance of FourIE in the cross-lingual transfer learning settings where the model is trained on English training data (source language) and tested on test data of the other languages in MEE. As can be seen, compared to performance on English test set, FourIE suffers from significant performance drops over different tasks and multilingual encoders when it is evaluated on other languages. It thus demonstrates inherent challenges of cross-lingual generalization for complete EE models that can be further studied with MEE. In addition, the performance loss due to cross-lingual testing varies across different target languages (e.g., 10.88\% loss for Spanish vs. 33.42\% loss for Japanese in EAE task). These variations can be attributed to different levels of divergence between languages (e.g., sentence structures and morphology) that hinder cross-lingual knowledge transfer for EE.

%We also report the performance of the best performing model, i.e., FourIE, in cross-lingual settings. To this end, we employ the training data of the English split, then we evaluate the trained model on the test sets of the other languages. The results are shown in Table \ref{tab:crosslingual_results}. As it is evident from this table, most languages suffer performance loss in this setting, which indicates the inherent differences for entity and event mentions in other languages. However, the performance loss across multiple language also varies (e.g., 10.88\% performance loss for Spanish vs 33.42\% for Japanese in EAE task). This divergence could be attributed to the different structures of the sentences and vocabularies shared across languages. 

\begin{table}[]
    \centering
    \resizebox{.48\textwidth}{!}{
    \begin{tabular}{l|ccc}
        Language & Entity & Event & Argument \\ \hline
        English \cite{devlin2019bert} & 70.21 & 73.18 & 66.19 \\
        Spanish \cite{CaneteCFP2020} & 67.29 & 65.14 & 60.13 \\
        Portuguese \cite{souza2020bertimbau} & 70.21 & 68.88 & 67.13 \\
        Polish \cite{polish2020bert} & 65.78 & 61.23 & 59.14 \\
        Turkish \cite{turkish2020bert} & 67.34 & 64.19 & 58.72 \\
    \end{tabular}
    }
    \caption{Test data performance (F1) of {\bf FourIE} in monolingual learning using available language-specific BERT models on MEE. The citations indicate the sources of the language-specific models.}
    \label{tab:bert_language_results}
\end{table}

\begin{table}[]
    \centering
    \resizebox{.48\textwidth}{!}{
    \begin{tabular}{l|ccc}
       Language & Entity & Event & Argument \\ \hline
       English \cite{liu2019roberta} & 70.32 & 72.28 & 69.19 \\ 
       Spanish \cite{spanish2020robert} & 70.23 & 61.34 & 60.28 \\
       Polish \cite{polish2020robert} & 68.12 & 60.89 & 60.34 \\
       Hindi \cite{hindi2020robert} & 64.91 & 59.09 & 60.38 \\
       Japanese \cite{japanese2020robert} & 69.72 & 60.45 & 71.45 
    \end{tabular}
    }
    \caption{Test data performance (F1) of {\bf FourIE} in monolingual learning using available language-specific RoBERTa models on MEE. The citations indicate the sources of the language-specific models.}
    \label{tab:roberta_results}
\end{table}

\noindent \textbf{Language-Specific Encoders}: To study the effectiveness of pre-trained language models as text encoders for EE models, we compare the performance of FourIE when the multilingual encoders mBERT or XLM-RoBERTa are replaced with comparable language-specific encoders (i.e., BERT-based models for mBERT and RoBERTa-based models for XLM-RoBERTa). Using publicly available pre-trained language models for our languages in MEE, Tables \ref{tab:bert_language_results} and \ref{tab:roberta_results} show the monolingual performance over test data of the languages for BERT-based and RoBERTa-based models (respectively). Comparing corresponding performance in Tables \ref{tab:mbert_results}, \ref{tab:xlmr_results}, \ref{tab:bert_language_results} and \ref{tab:roberta_results}, it is clear that language-specific language models all under-perform their multilingual counterparts over different EE tasks and languages, thus suggesting the benefits of multilingual data to train language model encoders to boost EE performance over different languages.

\begin{table}[]
    \centering
     \resizebox{.48\textwidth}{!}{
    \begin{tabular}{l|ccc|ccc}
        Language & \multicolumn{3}{c}{Trained on English} & \multicolumn{3}{c}{Trained on Polish} \\ \cline{2-7}
        & Entity & Event & Argument & Entity & Event & Argument \\ \hline
        Portuguese & 53.22 & 50.79 & 40.21 & 54.17 & 51.70 & 42.33 \\
        Spanish & 50.76 & 43.72 & 41.16 & 51.72 & 45.22 & 45.81 \\
        Turkish & 54.44 & 50.12 & 55.71 & 53.99 & 50.78 & 56.15 \\
        Hindi & 51.44 & 52.78 & 45.27 & 52.21 & 53.00 & 47.24 \\
        Japanese & 36.16 & 41.23 & 38.13 & 37.17 & 40.13 & 39.28 \\
        Korean & 43.72 & 40.08 & 37.29 & 42.08 & 39.78 & 37.10 \\
    \end{tabular}
    }
    \caption{Cross-lingual performance (F1) of \textbf{FourIE} with XLM-RoBERTa encoder  when it is trained on English or Polish training data, and tested on test data of the other languages in MEE. We use 3,500 random annotated segments from the training sets of English and Polish to train the model.}
    \label{tab:crosslingual_analysis}
\end{table}

%%%%%%In the proposed MEE dataset, we employ similar event definition and schema as provided in ACE 2005. As such, MEE is comparable to ACE 2005 entity and event extraction dataset. To study how challenging MEE is compared to ACE 2005, we conduct an analysis in which a model is trained on the ACE 2005, then it is evaluated on the test sets of the MEE for each languages. In particular, we utilize the standard data split from prior work \cite{nguyen2015event,chen2015event,wang2019adversarial} to prepare ACE 2005 dataset\footnote{It is note worthy that we remove the entity or event types in ACE 2005 that are not used in MEE}. For this experiments, we use FourIE with XLM-RoBERTa encoder which is shown to be the most effective model across all three tasks. The performance of the model on test sets of MEE, in terms of F1 score, is presented in Table \ref{tab:ace_analysis}. This table shows that the performance of the model trained on ACE 2005 significantly drops in comparison to training on MEE (i.e., see Table \ref{tab:XLM-RoBERTa_results}). This performance loss indicates the richer training signals provided to the model by MEE train set. 

\noindent {\bf Source Language Impact}: Finally, to study the impact of the source language for cross-lingual transfer learning for EE, we compare the performance of FourIE when either English or another comparable language is used as the source language to provide training data to train the model. In particular, we choose Polish as a comparable language for English as it has the same sentence structure (i.e., both languages have Subject-Verb-Object order) and entails similar density and type distributions for entity/event mentions as English. Table \ref{tab:crosslingual_analysis} shows the performance of the models when they are tested over test data of the other 6 languages in MEE. Here, to make it comparable, we use the same number of annotated segments (i.e., 3,500) sampled from training data of English and Polish to train the FourIE model. Interestingly, we find that Polish can lead to better performance for FourIE than English over a majority of task and target language pairs (i.e., over 4 languages for EMD and ED, and 5 languages for EAE). A possible explanation for this issue comes from richer event patterns that Polish might introduce to produce allow better cross-lingual generalization for EE than those for English. As such, this superior performance of Polish challenges the common practice of using English as the source language in cross-lingual transfer learning studies for EE and NLP. Future research can explore this direction to better understand the differences between languages to best select a source language to optimize performance over a target language for EE.

%Finally, we study the importance of the source language for cross-lingual setting. Specifically, we compare the effectiveness of the English as the source language with another comparable language. In this experiment, we choose Polish as comparable language as it has the same sentence structure as English, i.e., both languages are Subject-Verb-Object order, and also they show similar density and type distributions for entities and events. Since FourIE with XLM-RoBERTa encoder has shown the strongest performance in cross-lingual setting, we also employ this model to be trained on the train set of the source language. Table \ref{tab:crosslingual_analysis} shows the results, in terms of F1 score, for this experiment. As it is shown in this Table, the model trained on Polish generally outperforms the English version. Specifically, for NER and ED, in 4 languages and for EAE, in 5 languages, FourIE model achieves better performance if it is trained on Polish instead of English. This superiority challenges the common practice of using English as the source language in cross-lingual setting. This experiment could server as starting point to conduct more analysis on this direction. Specifically, the advantage of other languages compared to English could be attributed to the richer event and entity mention patterns in those languages which can be studied in future works.

\section{Related Works}

Due to its importance, various datasets have been recently developed for EE in different domains, including CySecED \cite{manduc2020introducing} (for cybersecurity domain), LitBank (for literacy) \cite{sims-etal-2019-literary}, MAVEN \cite{wang2020maven}, RAMS \cite{ebner2020multi}, and WikiEvents \cite{li-etal-2021-document} (for Wikipedia texts). However, these datasets are only annotated for English texts. There exist several multilingual datasets for EE, ACE \cite{walker05ace}, TAC KBP \cite{mitamura16overview,mitamura17event}, and TempEval-2 \cite{verhagen2010semeval}; however, such datasets only provide annotation for a handful of popular languages with limited number of event mentions and might not fully support all EE tasks (e.g., missing EAE in TAC KBP and TempEval-2).

%Our work introduces a much larger dataset for multilingual EE that covers 8 typologically different languages (5 of them are not explored in prior EE datasets) to enable more comprehensive development and evaluation for EE research.

Regarding model development, existing EE methods can be categorized into feature-based \cite{ahn2006stages,ji2008refining,liao2010filtered,hong2011using,li2013joint,yang2016joint} or deep learning \cite{chen2015event,nguyen2016joint,sha2018jointly,wang2019adversarial,lin2020oneie,pouran-ben-veyseh-etal-2021-unleash,pouran-ben-veyseh-etal-2021-modeling,liu-etal-2022-dynamic,pouran-ben-veyseh-nguyen-2022-word,nguyen-etal-2022-joint} methods. While most prior EE methods have been designed for one popular language, there have been growing interests in multilingual and cross-lingual learning for EE in recent work, featuring multilingual PLMs (i.e., mBERT and XLMR) as the key component for representation learning \cite{chen2009can,mhamdi2019contextualized,ahmad2020gate,nguyen2021crosslingual,huang2022multilingual,guzman-nateras-etal-2022-cross}. However, as such works only rely on existing multilingual EE datasets, their evaluation is limited to a few popular languages and fails to evaluate the generalization over many other languages.

\section{Conclusion}

We present a novel multilingual EE dataset, i.e., MEE, that covers 8 typologically different languages with more than 50K event mentions to support training of large deep learning models. MEE provides complete annotation for three EE sub-tasks, i.e., entity mention detection, event detection, and event argument extraction. To study the challenges in MEE, we conduct extensive analysis and experiments with different EE methods in the monolingual and cross-lingual learning settings. Our results demonstrate various challenges for EE in the multilingual settings that can be further pursued with MEE. In the future, we will extend our dataset to include more languages and tasks for IE.

%In this work, we present a novel multilingual event extraction dataset, i.e., MEE. Our dataset covers 8 typologically different languages, with more than 31k samples. MEE provides annotations for three sub-tasks, i.e., named entity recognition, event detection and event argument extraction, for language-specific articles in Wikipedia. In order to benefit from the rich prior research on event extraction, MEE follows ACE 2005 definition and schema for entities and events. To study the challenges in MEE, we conduct extensive analysis for entity recognition, event detection and event argument extraction in monolingual and cross-lingual settings. Our experiments and analysis corroborates the challenging nature of EE in multilingual setting. We publicly release MEE, to foster research in this direction. 

\section*{Limitations}

In this work we present a novel large-scale multilingual dataset for Event Extraction. As it is shown in the experiments, our dataset introduces many challenges that can inform future research on multilingual Event Extraction. However, there are still some limitations in the current work that can be improved in future research. First, cross-lingual transfer learning for EE is a challenging task that requires specifically designed models and methods. However, in this work, we have mainly focused on existing state-of-the-art EE models that are originally developed for the monolingual settings. As such, future work can study cross-lingual transfer models that are specifically designed to address the gaps between languages to better understand the challenges in our multilingual EE dataset. Second, in addition to data scarcity for multilingual EE, another challenge for this problem is the lack of resources for text encoding and processing in multiple languages. In particular, pre-trained language models and text processing tools might not be available for some languages (e.g., low-resource languages) that hinder dataset creation and model development efforts. As such, our work has not explored datasets and methods for low-resource languages for EE. In addition, as shown in our experiments, a majority of existing language-specific text encoders under-perform their multilingual counterparts for EE models. However, our work has not studied methods to improve such language-specific language models for EE. Finally, although our experiments empirically challenge English as the main source language for cross-lingual learning, we have not explored why other languages might be better options for the source language in this setting. Future research can perform more comprehensive analysis to shed light on this direction.

%In this work we presented a novel large-scale multilingual Event Extraction dataset. As it is shown in our experiments, this dataset is challenging and it can facilitate future research on multilingual Event Extraction research. However, there are some limitations in the current work which should be addressed in future research. First, cross-lingual EE is a challenging task that requires specific models and methods designed for this setting. However, in this work, we have employed the existing SOTA models that are mainly evaluated in monolingual setting. As such, it is necessary to study models and methods that are specifically designed to address language gaps (e.g., language-specific data augmentation). Second, in addition to data scarcity for multilingual EE, another challenge for this setting is the lack of resources and tools for text encoding and processing. As it is shown in our experiments, the majority of the existing language-specific text encoders under-perform their multilingual counterparts. In this work, we have not studied the influence of improving the language-specific pre-processing tools. Finally, although our experiments challenges English as the main source language for cross-lingual setting, due the the lack of space, in this work, we have not studied why other languages might be a better options for cross-lingual setting. In future, a more detailed analysis is required to shed light on this observation.

\section*{Acknowledgement}

This research has been supported by the Army Research Office (ARO) grant W911NF-21-1-0112 and the NSF grant CNS-1747798 to the IUCRC Center for Big Learning. This research is also based upon work supported by the Office of the Director of National Intelligence (ODNI), Intelligence Advanced Research Projects Activity (IARPA), via IARPA Contract No. 2019-19051600006 under the Better Extraction from Text Towards Enhanced Retrieval (BETTER) Program. The views and conclusions contained herein are those of the authors and should not be interpreted as necessarily representing the official policies, either expressed or implied, of ARO, ODNI, IARPA, the Department of Defense, or the U.S. Government. The U.S. Government is authorized to reproduce and distribute reprints for governmental purposes notwithstanding any copyright annotation therein. This document does not contain technology or technical data controlled under either the U.S. International Traffic in Arms Regulations or the U.S. Export Administration Regulations.

\bibliography{anthology,custom}
\bibliographystyle{acl_natbib}

\clearpage

\appendix

\end{document}